\pdfoutput=1
\documentclass[letterpaper, 10 pt, conference]{ieeeconf}  

\IEEEoverridecommandlockouts                              

\usepackage{epstopdf}

\usepackage{graphics} 
\usepackage{epsfig} 
\usepackage{amsmath} 
\usepackage{amssymb}  
\usepackage{enumerate}
\usepackage{float} 
\usepackage[linesnumbered,ruled]{algorithm2e} 
\usepackage{adjustbox}
\usepackage{subcaption} 
\usepackage{hyperref}
\usepackage[]{units}
\usepackage{bm}

\usepackage{amsfonts}

\usepackage{array}
\newcolumntype{P}[1]{>{\centering\arraybackslash}p{#1}}

\usepackage{tikz}
\usepackage{tikz-layers}
\usetikzlibrary{arrows.meta}
\usetikzlibrary{positioning, calc}
\usepackage{forest}
\def\nodeminsize{8}
\def\nodedistance{30}
\tikzset{
    roundnode/.style={ellipse, draw=black!60, fill=black!5, very thick, minimum size=\nodeminsize mm, node distance=\nodedistance mm, inner sep=1.5mm, font=\normalsize},
    rectanglenode/.style={rectangle, draw=black!60, fill=black!5, very thick, minimum size=\nodeminsize mm, node distance=\nodedistance mm, inner sep=1.5mm, font=\normalsize},
    blanknode/.style={rectangle, draw=black!0, fill=black!0, very thick, minimum size=\nodeminsize mm, node distance=\nodedistance mm, inner sep=1.5mm, font=\normalsize},
    edge from parent/.style={draw,->,black},
    line/.style={->},
    missionnode/.style={ellipse, draw=black!60, fill=black!5, very thick, minimum size=8 mm, node distance=40 mm, inner sep=1.5mm, font=\normalsize, align=center}, 
}

\title{\LARGE \bf
Reactive Multi-agent Coordination using Auction-based Task Allocation and Behavior Trees

}

\author{Niklas Dahlquist, Björn Lindqvist, Akshit Saradagi, and George Nikolakopoulos
\thanks{The authors are with the Robotics and AI Group, Department of Computer, Electrical and Space Engineering, Lule\r{a} University of Technology, Lule\r{a} SE-97187, Sweden}
\thanks{Corresponding Author's email: \texttt{Niklas.Dahlquist@ltu.se}}   
}

\begin{document}

\maketitle
\thispagestyle{empty}
\pagestyle{empty}

\begin{abstract}
This article presents an architecture for multi-agent task allocation and task execution, through the unification of a market-inspired task-auctioning system with Behavior Trees for managing and executing lower level behaviors. We consider the scenario with multi-stage tasks, such as 'pick and place', whose arrival times are not known a priori. In such a scenario, a coordinating architecture is expected to be reactive to newly arrived tasks and the resulting rerouting of agents should be dependent on the stage of completion of their current multi-stage tasks. In the novel architecture proposed in this article, a central auctioning system gathers bids (cost-estimates for completing currently available tasks) from all agents, and solves a combinatorial problem to optimally assign tasks to agents. For every agent, it's participation in the auctioning system and execution of an assigned multi-stage task is managed using behavior trees, which switch among several well-defined behaviors in response to changing scenarios. The auctioning system is run at a fixed rate, allowing for newly added tasks to be incorporated into the auctioning system, which makes the solution reactive and allows for the rerouting of some agents (subject to the states of the behavior trees). We demonstrate that the proposed architecture is especially well-suited for multi-stage tasks, where high costs are incurred when rerouting agents who have completed one or more stages of their current tasks.
The scalability analysis of the proposed architecture reveals that it scales well with the number of agents and number of tasks. 
The proposed framework is experimentally validated in multiple scenarios in a lab environment. A video of a demonstration can be viewed at: \href{https://youtu.be/ZdEkoOOlB2g}{https://youtu.be/ZdEkoOOlB2g}.

\end{abstract}

\section{Introduction}

\subsection{Motivation}
Teams of autonomous robots have made significant impact in various domains, such as, warehouse scenarios \cite{8901065, 9410352}, infrastructure inspection \cite{9210515}, search and rescue missions \cite{lindqvist2022multimodality, agha2021nebula} and logistics \cite{8392775}. In such domains, a team of multiple agents can speed up the task execution and enable execution of a wider range of tasks. Deploying multi-agent systems efficiently involves several complexities, with a major complexity lying in the management and coordination of multiple agents. This necessitates a coordination layer that quickly and adaptively decides how the currently available tasks should be distributed among a team of robots. In the case where the agents are acting in a dynamic environment, such as when sharing a workspace with human operators, the requirement of quickly reacting to changes in the environment greatly increases. Deploying a large number of robots in a small and constrained space also increases the probability of inter-agent collisions \cite{7081368}. Additionally, in many cases, the arrival of tasks is not known a priori or tasks are introduced during run-time, thus requiring quick reconfiguration for optimal behavior.
\begin{figure}
    \centering
    \includegraphics[width = 0.95\columnwidth]{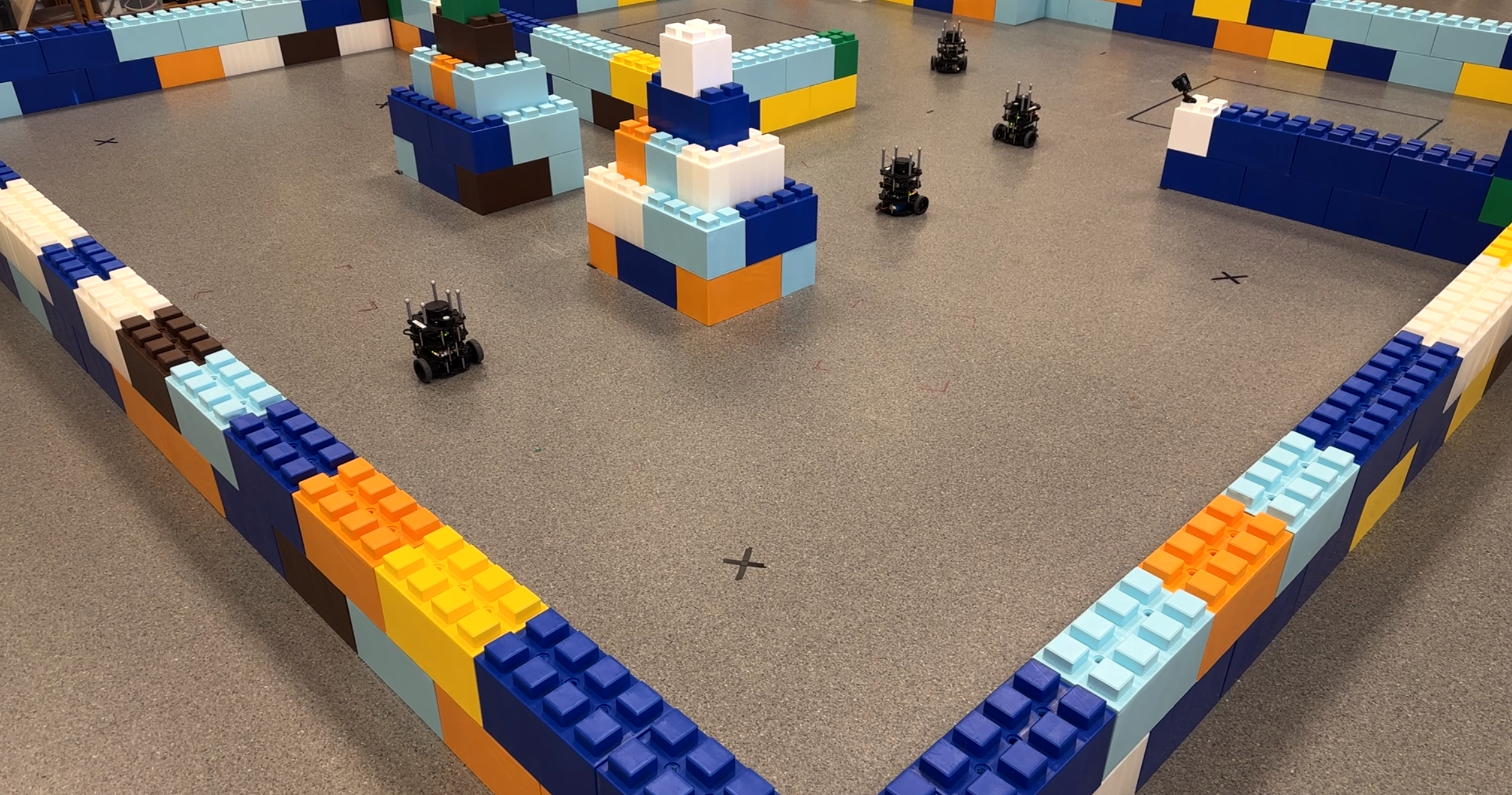}
    \caption{Overview of an experimental setup where a team of multiple agents (TurtleBot3) reactively accomplish multi-stage tasks in a lab environment.}
    \label{fig:lab_example}
\end{figure}
\subsection{Background}
Research in multi-agent systems has received a lot of attention in the field of robotics in the recent years \cite{9029554, 9220149}, while some well-studied problems include vehicle routing and coalition formation \cite{https://doi.org/10.48550/arxiv.2207.09650}. The problem of task allocation, where agents are assigned to available tasks, is encountered often in several applications. Various approaches for solving task-allocation problems can be found in the related literature; including those based on optimization, where a global objective function is minimized/maximized, which usually requires global information about the state of the multi-agent system; deterministic approaches that generate optimal solutions but may be too time consuming \cite{6225234}; heuristic approaches, such as \cite{heuristics_task_allocation}; and market-inspired approaches, where tasks are treated as objects that can be transferred between agents, either through a centralized coordinator or through consensus \cite{9029554, 5980500}. It is common in this literature to assume, as in \cite{6213575}, that all tasks to be completed are known in advance.

In this article, we consider the scenario with multi-stage tasks, such as 'pick and place', whose arrival times are not known a priori. In such a scenario, a coordinating architecture is expected to be reactive to newly arrived tasks and the resulting rerouting of agents should be dependent on the stage of completion of their current multi-stage tasks, due to the high costs incurred when rerouting agents who have completed one or more stages of their current tasks. Our review of literature revealed that a scalable reactive solution to the scenario of multi-stage tasks, whose arrival times is not known a priori, has not been presented before and this article addresses this problem using Behavior Trees for executing multi-stage costs. The concept of Behavior trees, which has origins in computer science, has seen an increased adoption in the recent years in robotics and automation research \cite{DBLP:journals/corr/abs-2005-05842}
 \cite{9448466}. This is due to their ability to create modular and reconfigurable intelligence for autonomous agents. Behavior trees also generalize other popular control structures, such as finite state machines and decision trees \cite{BT_introduction}, thus increasing its utility as a flexible and versatile framework for automation.


\subsection{Contributions}
With respect to the aforementioned state-of-the-art, the contributions of this article can be summarized as follows.
\begin{enumerate}
\item This work presents a novel scalable and reactive multi-agent architecture for the case of multi-stage tasks whose arrival times are not known a priori. The architecture composes of a light central market-inspired auctioning system that collect bids from the agents for the available tasks, solves an optimization problem and assigns tasks to agents. The agents have the autonomy of computing the costs, taking part in the auction and in executing the assigned tasks in a complex unstructured environment using modular behavior trees. 
\item The autonomy of the agents significantly reduces the complexity of the central allocation and allows for low-cost execution of the auctioning system at high rates. This makes the architecture reactive towards newly added tasks. As the costs for completing the tasks are computed locally at each agent, the task allocation layer has no need for full global knowledge. Multi-agent coordination is achieved with very little information exchange between the agents and the task-allocator.
\item The proposed architecture is especially well-suited for reallocation of multi-stage costs, especially when there are high costs for abandoning tasks with one or more completed stages. The behavior tree involved in the execution of multi-stage tasks controls the participation of an agent in the auction, depending on the state of execution of the current task. Such an interaction between task allocation (central planning) and behavior trees (local task execution) has not be proposed before, to the best of the authors' knowledge.
\item We present an experimental validation of the architecture in a laboratory environment (shown in Figure \ref{fig:lab_example}). 
In this approach, a risk-aware path planning is used by the agents to compute costs and distributed nonlinear model predictive control is used for path-tracking and inter-agent collision avoidance. The results demonstrate complete autonomous multi-agent execution of multi-stage tasks.

\end{enumerate}

\section{Problem Formulation}
The set of tasks available for allocation, at a specific time instant \(t\), is denoted by \( \mathfrak{T}_t = \{ T_1, \dots, T_{n_t} \} \), where \(n_t \in \mathbb{N}\) is the number of tasks available at time $t$. A team of agents \(\mathfrak{R} = \{ R_1, \dots, R_{n_a} \}\) are available for executing the tasks, where \(n_a \in \mathbb{N}\) is the number of agents. 
It is also assumed that the times at which the tasks are introduced is not known a priori. Each task in the set $\mathfrak{T}_t$ is a multi-stage task and to be completed by one agent. Without loss of generality and with the aim of simplifying the notations, we consider two-stage "pick and deliver" tasks throughout this article. In accomplishing such a task, an agent first moves to a specific location to pick-up an object and then moves to a target drop-off location.
To precisely define the scenario under consideration, we make the following practical assumptions.

\emph{Assumption 1:} Every agent \(R_k\) is able to complete an assigned task \(T_j\) without failing or breaking down. Abandoning a task without completing it is not allowed, except when the centralized auction system reassigns an agent to a different task.

\emph{Assumption 2:} There exist reliable communication links between all agents in \(\mathfrak{R}\) and a centralized computing system. The communication links are used to share information about available tasks, costs for completing the tasks, communicating task assignment and for distributed inter-agent collision avoidance.

\emph{Assumption 3:} All agents in \(\mathfrak{R}\) are equipped with a local static map of the environment. This map is used for risk-aware path planning to estimate costs and for collision-free navigation to desired locations.




\emph{Problem Definition:}
Using the notations, definitions and assumptions made so far, the problem considered in this article can be stated as follows.

We are considering the scenario where the multi-stage tasks to be completed are not predefined and the times that tasks are made available is unknown beforehand. For this scenario, the problem of interest is to design a scalable and reactive multi-agent coordination architecture that decides how the available tasks \(\mathfrak{T}_t\) should be allocated among the agents \(\mathfrak{R}\) and drives the agents to complete the tasks. To decide the optimal task allocation, the architecture must repeatedly minimize the total cost incurred in completing the currently assigned tasks. The optimization problem to be solved for optimal task allocation is defined as:
\begin{equation}
    \min_{x_{k, j} \in \{0, 1\}} \sum_{k,j\in E} c_{k,j} \cdot x_{k,j}
\end{equation}
where \(x_{k,j}\in\{0,1\}\) is an assignment variable defining if agent \(R_k\) is assigned to task \(T_j\), \(E=\{1,\ldots,n_a\}\times\{1,\ldots,n_t\}\) and \(c_{k,j}\) is the cost for agent \(R_k\) to complete task \(T_j\). The following constraints are also imposed:
\begin{equation}
    \begin{aligned}
        \sum_{\{k \mid R_k \in \mathfrak{R}\}} x_{k,j} &\leq 1, \text{ }  \forall j \in \{1, \ldots, n_t\} \\
        \sum_{\{j \mid T_j \in \mathfrak{T}_t\}} x_{k,j} &\leq 1, \text{ }  \forall k \in \{1, \ldots, n_a\} \\
    \end{aligned}
    \label{constraints}
\end{equation}
to ensure that not more than one agent is assigned to any task and that a specific task is not assigned to multiple agents.

\section{Methodology}
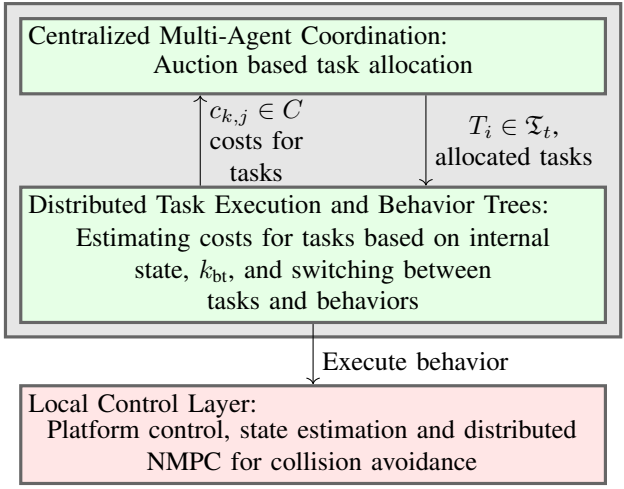
\begin{figure}[h]
    \centering
    \begin{tikzpicture}[
        align=center,
        ]
        \node[draw=black!60, fill=green!10, very thick, minimum width=0.9\columnwidth, minimum height = 10mm] (layer_1_text) at (0 mm, 0 mm) {};
        \node[minimum width=0.9\columnwidth, minimum height = 10mm] (layer_1_text_new) at (0 mm, -1 mm) {Auction based task allocation};

        \node[anchor=north west] (layer_1) at (layer_1_text.north west) {Centralized Multi-Agent Coordination:};
        \node[anchor=north, fill=green!10, draw=black!60, very thick, minimum width=0.9\columnwidth, minimum height = 18mm] (layer_2_text) at ($(layer_1_text.south) + (0, -12mm)$) {};
        \node[anchor=north, minimum width=0.9\columnwidth, minimum height = 18mm] (layer_2_text_new) at ($(layer_1_text.south) + (0, -14mm)$) {Estimating costs for tasks based on internal\\state, \(k_\text{bt}\), and switching between\\ tasks and behaviors};
        \node[anchor=north west] (layer_2) at (layer_2_text.north west) {Distributed Task Execution and Behavior Trees:};

        \node[anchor=north west, draw=black!60, fill=red!10, very thick, minimum width=0.9\columnwidth, minimum height = 13mm] (layer_3_text) at ($(layer_2_text.south west) + (0, -8mm)$) {};
        \node[anchor=north west, minimum width=0.9\columnwidth, minimum height = 13mm] (layer_3_text_new) at ($(layer_2_text.south west) + (0, -9.5mm)$) {Platform control, state estimation and distributed\\NMPC for collision avoidance};
        \node[anchor=north west] (layer_3) at (layer_3_text.north west) {Local Control Layer:};
        
        \draw[line] ($ (layer_1_text.south) + (15mm, 0) $) -- node[anchor=west] {$T_i \in \mathfrak{T}_t$,\\allocated tasks} ($ (layer_2_text.north) + (15mm, 0) $);
        \draw[line] ($ (layer_2_text.north) + (-15mm, 0) $) -- node[anchor=west] {$c_{k,j} \in C$\\costs for\\tasks} ($ (layer_1_text.south) + (-15mm, 0) $);

        \begin{scope}[on behind layer]
            \node[anchor=north, draw=black!60, fill=black!10, very thick, text width=0.92\columnwidth, text height = 42mm] (box) at ($(layer_1_text.north) + (0, 2mm)$) {};
        \end{scope}

        \draw[line] ($ (layer_2_text.south) + (-0mm, 0) $) -- node[anchor=west, yshift=-1mm] {Execute behavior
        } ($ (layer_3_text.north) + (-0mm, 0) $);

    \end{tikzpicture}
    \caption{Overview of the overall structure of the proposed architecture combining an auction system for reactive task allocation with behavior trees for local control and cost estimation and incorporating collision avoidance through distributed NMPC.}
    \label{fig:overview_structure}
\end{figure}

In this section, we present a short overview of our proposed solution to the problem described in the previous section. The proposed architecture for multi-agent coordination is structured as in Fig. \ref{fig:overview_structure}. 

A centralized auctioning system keeps tracks of the arrival of new tasks and announces the currently available tasks to the agents. The agents then compute the costs, depending on their state and taking into account the stage of execution of the current multi-stage task. The auctioning system gathers the costs and optimally allocates tasks to agents. The auctioning system is run at a reasonably high rate to reactively incorporate newly introduced tasks. Allocated tasks are then completed using behavior trees. A grid based risk-aware path planning is used to both estimate the costs related to the tasks and to complete the allocated tasks. Inter-agent collision avoidance is handled through a distributed NMPC scheme.

In the following sections, we present detailed descriptions of the modules constituting the proposed architecture.

\section{Auction-based Task Allocation}

\begin{figure}[t]
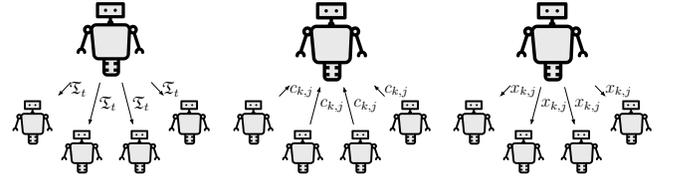

    \centering
    \begin{subfigure}[t]{0.32\columnwidth}
        \centering
        \input{figures/tex/auction_fig_1}
        \caption{Announcement stage. Available tasks \(\mathfrak{T}_t\) are announced to all agents.}
    \end{subfigure}
    \hfill
    \begin{subfigure}[t]{0.32\columnwidth}
        \centering
        \input{figures/tex/auction_fig_2}
        \caption{Bidding stage. Agents calculate costs \(c_{k, j}\) and return them to the auctioneer.}
    \end{subfigure}
    \hfill
    \begin{subfigure}[t]{0.32\columnwidth}
        \centering
        \input{figures/tex/auction_fig_3}
        \caption{Task allocation stage. The auction system optimizes how to assign tasks and broadcasts \(x_{k, j}\).}
    \end{subfigure}
    \caption{An illustration of the different stages of the auction system.}
    \label{fig:auction_summary}
\end{figure}

The market-inspired auction system comprises of three stages: 
1) Announcement stage: The auctioning system announces the available tasks \(\mathfrak{T}_t\) to the agents. 
2) Bidding stage: The agents compute the costs (\(c_{k,j}\)) for accomplishing each task and bids are submitted to the auctioning system.
3) Task allocation stage: The task allocator solves a combinatorial optimization problem to decide how the tasks \(\mathfrak{T}_t\) should be distributed among the agents \(\mathfrak{R}\). The allocated tasks are then announced to the agents.
Fig. \ref{fig:auction_summary} summarises the three stages. Because the cost computations are performed locally by the agents, the central allocator has no need for the global information of the multi-agent system and this significantly reduced the complexity of the combinatorial task allocation problem. Because of the modularity and simplicity of the auctioning system, the three constituent steps can be run at a very high-rate. In this work, we intentionally introduce small delays to set a reasonable rate, so that the auction system is reactive enough to quickly accommodate newly introduced tasks into the auction.


\subsection{Formulation of the task allocation problem : Integer Linear Program }
\label{sec:auction_ilp}
The problem of deciding what task should be allocated to which agent, in a way that minimizes the total cost and assigns the maximum number of tasks, can be modelled as an undirected bipartite graph \(G = (\mathfrak{R}, \mathfrak{T}_t, E)\), where \(\mathfrak{R}\) and \(\mathfrak{T}_t\) respectively represent the agents and the available tasks and form the disjoint nodes of the bipartite graph; and \(E=\{1,\ldots,n_a\}\times\{1,\ldots,n_t\}\) is the set of edges of the bipartite graph. Each edge in $E$ is associate with a weight $c_{i,j}$ which represent the cost of a potential match between a task and an agent.
As an example, the edge between agent \(R_k\) and task \(T_j\) represents the cost for agent \(R_k\) to perform task \(T_j\). To fit the above setting into an auction framework, the costs
\begin{equation}
    c_{k,j} = \text{cost of assigning agent \(R_k\) to task \(T_j\)}
\end{equation}
are converted, in a way that preserves the inverse order of the costs \(c_{k,j}\), to profits
\begin{equation}
    \rho_{k,j} = \text{profit of assigning agent \(R_k\) to task \(T_j\)}
    \label{inversion}
\end{equation}
and an assignment variable
\begin{equation}
    x_{k,j} = 
    \begin{cases}
        1,  \text{ if \(R_k\) is matched to \(T_j\)}\\
        0, \text{ otherwise}
    \end{cases}, (k,j) \in E
\end{equation}
is introduced as an optimization variable. The auctioning system looks to maximize the sum of the profits (equivalent to minimizing the sum of costs) and at the same time looks to assign maximum number of tasks to agents. The objective function to maximize is then formulated as
\begin{equation}
    \max_{x_{i,j}} \sum_{(k,j) \in E} \rho_{k,j} \cdot x_{k,j}
    \label{eq:auction_optimization}
\end{equation}
with the constraints
\begin{equation}
    \begin{aligned}
        \sum_{\{k \mid R_k \in \mathfrak{R}\}} x_{k,j} &\leq 1, \text{ }  \forall j \in \{1, \ldots, n_t\} \\
        \sum_{\{j \mid T_j \in \mathfrak{T}_t\}} x_{k,j} &\leq 1, \text{ }  \forall k \in \{1, \ldots, n_a\} \\
    \end{aligned}
    \label{constraints}
\end{equation}
posed to constrain the solution so that at most one task is allocated to each agent and to make sure that any task is not allocated to more than one agent.


It should be noted that using this optimization formulation is very flexible and allows, for example, to include tasks with different priorities. The terms \(\rho_{k,j}\) in equation \eqref{eq:auction_optimization} associated with different tasks can be multiplied with scaling constants indicative of the relative priorities of different tasks.


\subsection{Numerical Study of the Scaling of the Architecture}\label{numerical_study}
\begin{figure}
    \centering
    \includegraphics[width = 0.8\columnwidth]{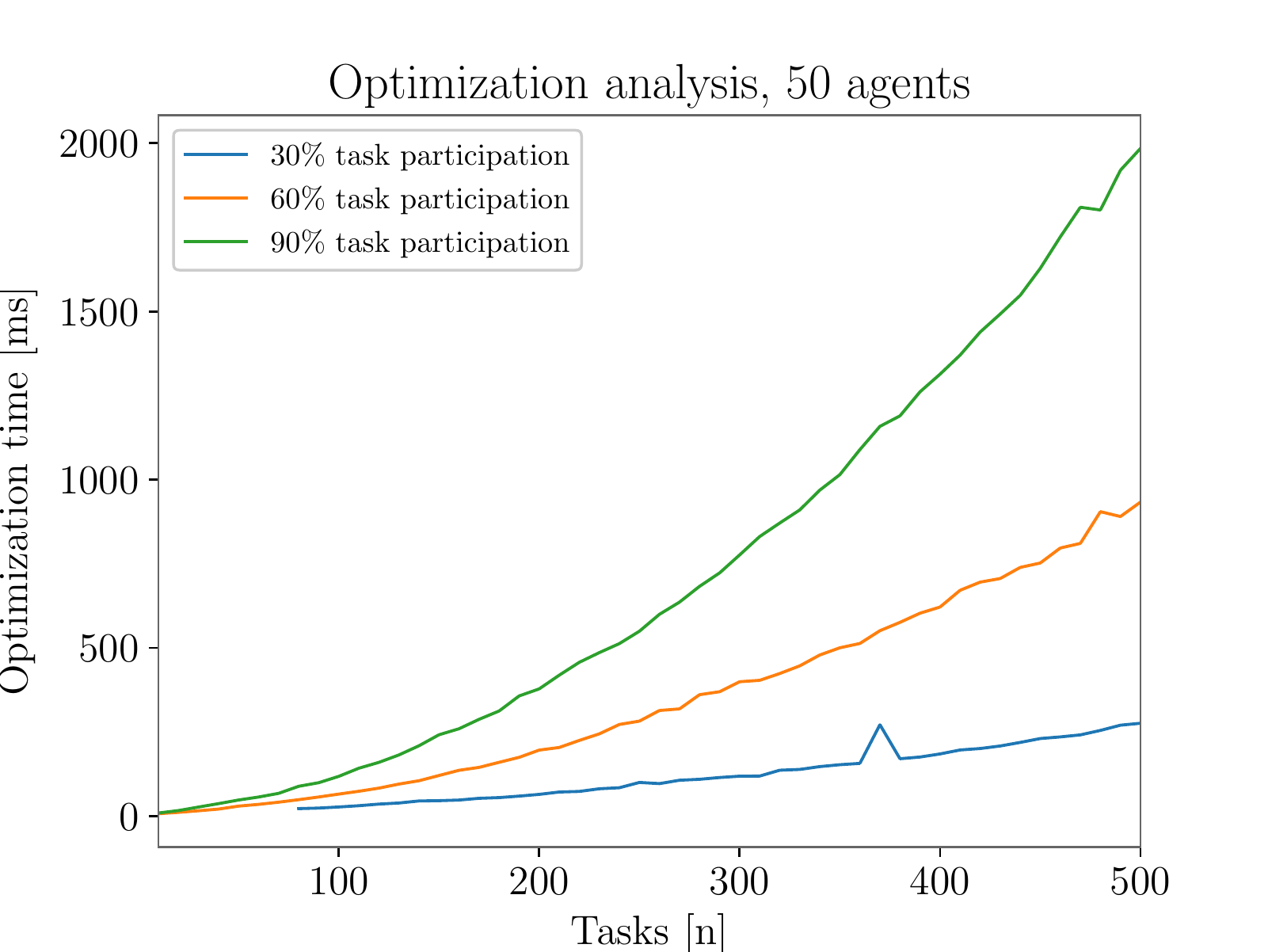}
    \caption{
    The time to solve the task allocation optimization problem with increasing number of agents, while the number of tasks is fixed. The three curves in the plot show the scaling when agents only bid for a percentage (or for a subset) of the available tasks.
    }
    \label{fig:optimization_constant_agents}
\end{figure}

\begin{figure}
    \centering
    \includegraphics[width = 0.8\columnwidth]{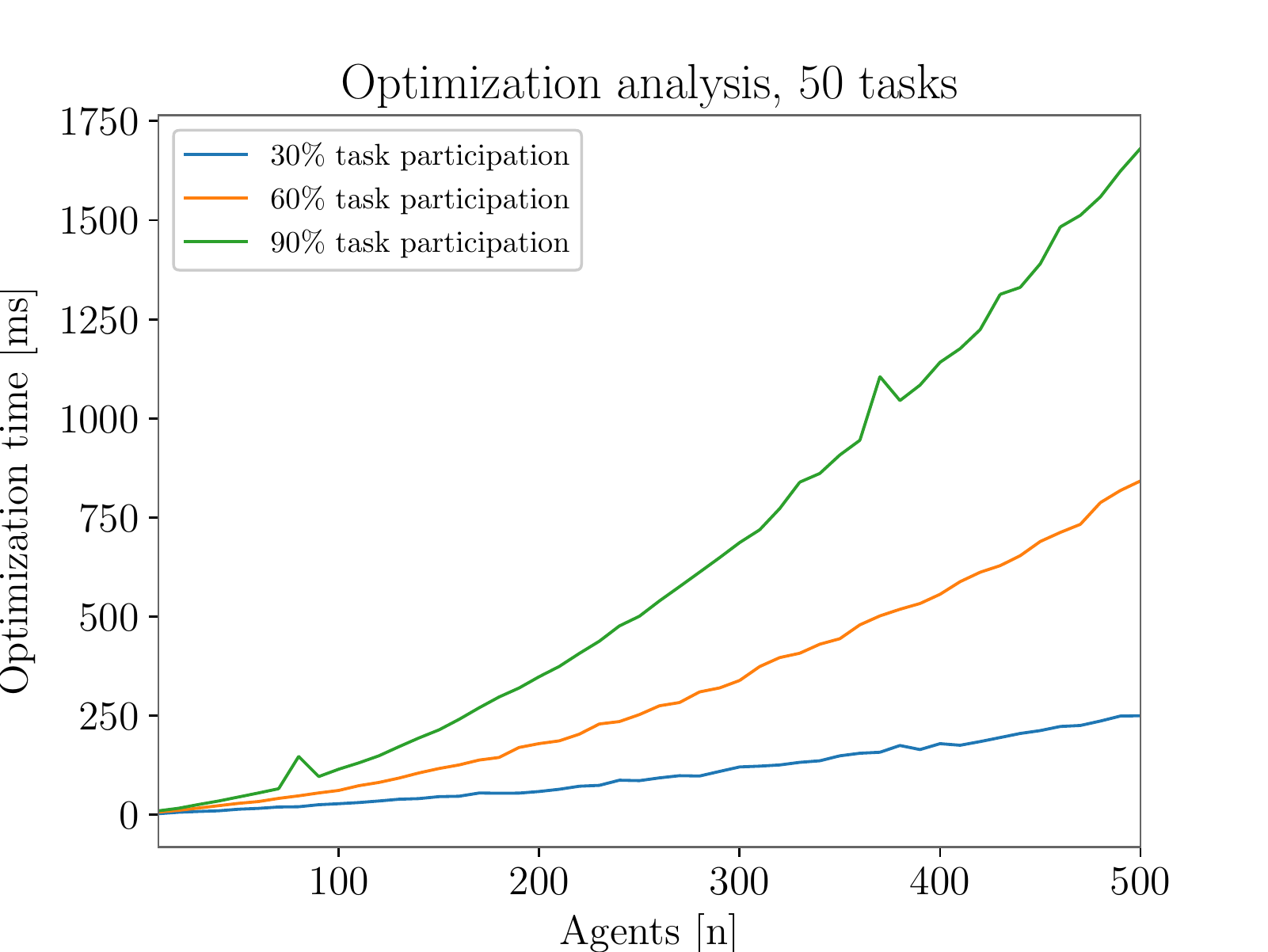}
    \caption{The time to solve the task allocation optimization problem with increasing number of tasks, while the number of agents is fixed. The three curves in the plot show the scaling when agents only bid for a percentage (or for a subset) of the available tasks. 
    .}
    \label{fig:optimization_constant_tasks}
\end{figure}

To assess how the proposed optimization problem would scale with the number of agents and with the number of tasks, we perform a numerical study through simulations in multiple scenarios. In Fig. \ref{fig:optimization_constant_agents} and Fig. \ref{fig:optimization_constant_tasks}, we plot the computation time to solve the task-allocation optimization problem for different task participation percentages (how many of the available tasks each agent bids for), by successively increasing the number of available tasks and the number of agents. The integer linear program (ILP) is defined using a software suite for optimization named OR-Tools \cite{ortools} and solved using SCIP \cite{GamrathEtal2020OO}, a constrained integer programming solver suite.

The simulation results reveal that, if the agents place bids for a subset of the available tasks, the time to solve the problem scales more favourably, compared to the case where all agents compete for all available tasks. Consider the case in Fig. \ref{fig:optimization_constant_agents}, with 50 agents and 400 tasks. The time to solve the optimization problem is reduced from \(1450\) ms to only \(180\) ms when the task participation decreases from 90\% to 30\%. Allowing agents to only submit bids for a subset of the tasks also decreases the time to compute costs locally at the agents. This shows that the auctioning system can scale to hundreds of tasks/agents and still be reactive enough to be practically usable.

When analysing the scalability of the proposed architecture, it is also important to consider the time taken to perform local computations at the agents, as this influences the reactivity of the auctioning system to newly introduced tasks. It should be noted that increasing the number of agents, while fixing the number of tasks will not impact the local computations at the agents associated with estimating costs for the tasks. Increasing the number of available tasks will, on the other hand, linearly increase the local cost computation for every agent.




\section{Task Execution and Behavior Trees}
In the proposed architecture, each agent uses behavior trees to execute the multi-stage costs and to interact with the central auctioning system. 
\subsection{Task Behavior Trees}
A behavior tree is a logical framework to decide how an agent switches dynamically between a number of behaviors. 
A behavior tree comprises of a number of nodes connected as a directed acyclic graph, where the internal nodes are called \emph{control nodes} and leaf nodes are called \emph{execution nodes}. The behavior tree is executed by "ticking" the root node. The "tick" is then passed down the tree by the control nodes until an execution node receives it. More details about the different nodes and the method of executing a behavior tree are described in \cite{BT_introduction}.

In a reactive task allocation scenario, the agents must keep track of the state of completion of the current multi-stage task, to know if it is possible to abandon the current task with low costs and to indicate availability to the auctioning system for reallocation. 
In the middle of a task execution, the condition nodes are used to track the current state through the variable ($k_{\text{bt}}$). The costs submitted by the agents in a new round of auctioning (\(c_{k,j}\)) are dependent on the current state of the behavior tree. 

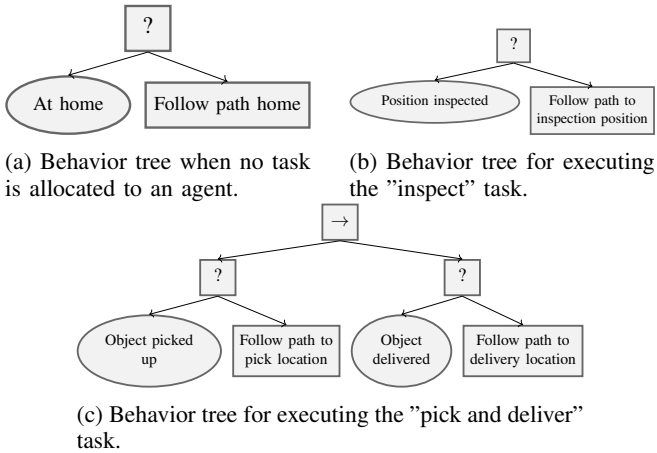
\begin{figure}[t]
    \centering
    \begin{subfigure}[t]{0.47\columnwidth}
        \centering
    \begin{adjustbox}{max width = \columnwidth}
        \begin{forest}
        for tree={align=center, edge={->, draw}, parent anchor=south,child anchor=north}
        [\large ?, rectanglenode
            [At home, roundnode]
            [Follow path home, rectanglenode]
        ]
        \end{forest}
    \end{adjustbox}
        \caption{Behavior tree when no task is allocated to an agent.}
        \label{fig:bt_no_task}
    \end{subfigure}
    \hfill
    \begin{subfigure}[t]{0.47\columnwidth}
        \centering
    \begin{adjustbox}{max width = \columnwidth}
        \begin{forest}
        for tree={align=center, edge={->, draw}, parent anchor=south,child anchor=north}
            [\large ?, rectanglenode
                [Position inspected, roundnode]
                [Follow path to\\inspection position, rectanglenode]
            ]
        \end{forest}
    \end{adjustbox}
        \caption{Behavior tree for executing the "inspect" task.}
        \label{fig:bt_inspect}
    \end{subfigure}
    \hfill
    \begin{subfigure}[t]{0.78\columnwidth}
        \centering
    \begin{adjustbox}{max width = \columnwidth}
        \begin{forest}
        for tree={align=center, edge={->, draw}, parent anchor=south,child anchor=north}
        [\large \(\rightarrow\), rectanglenode
            [\large ?, rectanglenode
                [Object picked\\up , roundnode]
                [Follow path to\\pick location, rectanglenode]
            ]
            [\large ?, rectanglenode
                [Object\\delivered, roundnode]
                [Follow path to\\delivery location, rectanglenode]
            ]
        ]
        \end{forest}
    \end{adjustbox}
        \caption{Behavior tree for executing the "pick and deliver" task.}
        \label{fig:bt_pick_place}
    \end{subfigure}
    \caption{The behavior trees used in the evaluation scenarios. It should be noted that, in (c), once the condition "object picked up" is satisfied, \(k_\text{bt}\) is set to zero.}
    \label{fig:bt_all}
\end{figure}

In this work, there are three different tasks considered for each agent, with the associated behavior trees shown in Fig. \ref{fig:bt_all}. 
The first behavior tree, shown in Fig \ref{fig:bt_no_task}, makes sure that the agent moves to its home position when it is not allocated to a task. The second, shown in Fig. \ref{fig:bt_inspect}, allows the agent to perform a one-stage "inspection" task by moving to the requested inspection position. Finally, the behavior tree for completing the two-stage "pick and deliver" task, shown in Fig. \ref{fig:bt_pick_place}, drives the agent to first move to a pick up location and when that stage is completed, drives the agent to move to the associated delivery location.

\subsection{Cost Estimation and Path Planning}

The computation of costs \(c_{k,j}\), for the available tasks \(T_j\) are done locally by every agent \(R_k\). The cost \(c_{k,j}\) submitted by an agent not only depends on the actual costs for completing a task but also on the state of execution of the current task. 
In this article, the costs \(c_{k,j}\) are computed in the following manner:
\begin{equation}
    c_{k,j} = f_\text{cost}(\text{task}, \text{ state}) \cdot k_\text{bt}
    \label{cost_function}
\end{equation}
where $f_\text{cost}(\text{task}, \text{ state})$ is the actual cost incurred by an agent, dependent on the type of task and the state of the agent and \(k_\text{bt}\) is a factor that depends on the stage of execution of the current task, derived from the state of the current behavior tree. The factor \(k_\text{bt}\) is set to one, if the stage of execution of the current task is such that very low costs are incurred if the agent is reallocated to another task. The value of \(k_\text{bt}\) is set to zero after a critical stage of a task execution has been completed, following which the cost of reallocation is high. This is to make sure that the auctioning system, after performing the inversion as in equation \eqref{inversion}, assigns very high profit to the current task and makes sure that an agent is not reallocated to another task.  For the "pick and place" tasks, shown in Fig. \ref{fig:bt_all}, this critical stage is when an agent picks up an object. Following this, \(k_\text{bt}\) is set to zero and the agent cannot be allocated to another task.

\subsubsection{Path Planning}
In the context of a delivery or "go-to-point" mission, the conventional cost considered in the task allocation framework is the length of the shortest path from the agent location to the task-goals. In a cluttered environment, the shortest robot-safe path to the goal could be considered, where the size of the robotic platform is taken into account in the planning process, for a more realistic evaluation of how the agent should reach it's goal safely. In this article, we use a grid-search algorithm DSP~\cite{karlsson2021d} as the path planner to compute the cost $f_\text{cost}(\text{task}, \text{ state})$ in equation \eqref{cost_function}. This algorithm incorporates a risk-layer to avoid entering high-risk areas (defined by being close to an occupied cell) if it is not absolutely necessary to reach the goal. DSP plans a path $P$ from robot position $\hat{p}$ to the task goal $p_g$ such that the
cost $\Gamma = \Gamma_\mathrm{dist} + \Gamma_\mathrm{risk}$, where \(\Gamma_\text{risk}\) represents the risk of moving close to obstacles and \(\Gamma_\text{dist}\) is the total distance to the goal position, is minimized. 
In the experimental results presented in section \ref{sec:Exp}, DSP is used both for generating safe paths and for estimating the distance to various positions.

\subsubsection{Task Specific Cost Functions}
The cost functions used for the tasks "inspect" and "pick and deliver" are chosen as
\begin{equation}
    f_\text{inspect} = \Gamma_\text{1} \cdot k_\text{bt}
\end{equation}
and \begin{equation}
    f_\text{pick and deliver} = \Gamma_\text{2} \cdot k_\text{bt}
\end{equation}
respectively, where \(\Gamma_\text{1}\) is the cost of the risk-aware path planned from the current position to the inspection position and \(\Gamma_\text{2}\) is the sum of the cost of the risk-aware path planned from the current position to the pick up position and the cost of the risk-aware path from the pick up position to the delivery position i.e. the total cost of be path for completing the two-stage task. For the pick and deliver tasks, the factor \(k_\text{bt}\) is set to one if the "pick" stage has not been completed and is set to zero if the "pick" stage has been completed, which reduces the cost for its current task and guarantees that the agent is not reallocated to another task.

\section{NMPC with integrated Collision Avoidance}
A critical aspect of any multi-agent coordination problem is to ensure that no collisions occur among agents during mission execution. Towards ensuring inter-agent safety during task execution, this article incorporates a nonlinear model predictive controller (NMPC) based on the preliminary work in \cite{lindqvist2021scalable}, but applied to ground robots as opposed to UAVs. The formulation allows for integration of the inter-agent collision avoidance constraints into a model-based predictive control layer. In this framework, each agent shares predicted NMPC trajectories with other agents and we form set-exclusion constraints on the position-space, based on the shared predicted trajectories. The TurtleBots considered for experimentation in this work, accept high-level actuation commands $u = [u_{v}, u_\omega]$, where $u_{v}$ is a forward/backward velocity command and $u_\omega$ is an angular velocity command. We use the nonlinear kinematic model of the TurtleBots:
\begin{subequations}
\label{eq:turtlekinematic}
\begin{align}
        \dot{p}_x(t) &= \cos{\psi(t)}u_{v}(t) \notag\\ 
        \dot{p}_y(t) &= \sin{\psi(t)}u_{v}(t) \notag \\
        \dot{\psi}(t) &= u_{\omega}(t) \tag{\ref{eq:turtlekinematic}}
\end{align}
\end{subequations}
with $p$ denoting the position coordinate in the global-frame and $\psi$ denoting the heading angle state. The state of the TurtleBots is denoted as $x = [p_x, p_y, \psi]$. We note that the robot velocity actuation $u_v$ is in its body-frame. We are interested in tracking reference way-points $x_\mathrm{ref}$ while minimizing the actuation effort and the change in actuation for consecutive time steps (so as to minimize oscillations). The objective function $J(\bm{x}_{k}, \bm{u}_{k}; u_{k-1\mid k})$ is as in our previous work \cite{lindqvist2021scalable} that includes penalties on the state deviation from the reference, inputs costs, input rate costs (deviations on consecutive control inputs), and terminal state costs. To include collision avoidance constraints, we follow the approach in \cite{sathya2018embedded} and impose a set-exclusion constraint~\cite{hermans2021penalty} on the robot position states, modelling the obstacle as a circular area with center $p_\mathrm{obs} = [p_\mathrm{obs,x}, p_\mathrm{obs,y}]$ and radius $r_\mathrm{obs}$. We can write this as an equality constraint using the $max(a,0) = [a]_+$ operator, such that the constraint is violated ($\leq 0$) if the obstacle and the robot are closer by distance than $r_{\mathrm{obs}}$:
\begin{multline}\label{eq:circleconstraint} 
    C_\mathrm{circle}(p, p_\mathrm{obs}, r_\mathrm{obs}) = [r_{\mathrm{obs}}^2 - (p_x{-} p_{\mathrm{obs,x}})^2 \\
    - (p_y{-}p_{\mathrm{obs,y}})^2]_+ = 0.
\end{multline}
The constraint is posed for all predicted time steps $j = 0 \ldots N$ (with $N$ denoting the prediction horizon), and as such we can include the concept of a moving obstacle by computing the predicted obstacle positions in the global frame along the horizon as $\bm{p}_\mathrm{obs}\in \mathbb{R}^{3\times N}$, and feeding them to the optimizer. Since we discretize the model \eqref{eq:turtlekinematic} with a sampling time $T_S$ as $x_{k+1} = \zeta(x_k, u_k)$, it should be clear that from every agent's trajectory $\bm{u}_k^i$ and measured state $\hat{x}^i$, we can form such an obstacle trajectory $\bm{p}_\mathrm{obs}^{i}$. This results in a distributed NMPC scheme where each agent solves its own NMPC problem, but the NMPC trajectories are shared among agents in the system, in this case for predictive collision avoidance. As the NMPC is used with a physical system, we also impose constraints on the control inputs as $u_\mathrm{min}\leq u_{k+j|k} \leq u_\mathrm{max}$, since the robot has a maximum actuation level before the motors saturate. The solving of the resulting NMPC problem follows the previous work \cite{lindqvist2020nonlinear}, using the Optimization Engine\cite{sopasakis2020open} where PANOC \cite{stella2017simple} is combined with a quadratic Penalty Method \cite{hermans2021penalty}.
The proposed NMPC is used in the following experiments to enable tracking global reference paths from DSP, while proactively preventing collisions among agents in the system. The NMPC runs with a sampling time of $\unit[0.1]{s}$ and a horizon of $N = 50$, implying a $\unit[5]{s}$ prediction horizon.
 
\section{Experimental Setup and Results}\label{sec:Exp}

For experimentally evaluating the proposed architecture, three scenarios were considered. The evaluations were performed at the Robotics Team's laboratory at Lule{\aa} University of Technology using TurtleBots\footnote{https://emanual.robotis.com/docs/en/platform/turtlebot3/overview/} as the agents. 

In all scenarios, a static map was used by all agents for path planning and cost estimation. Prior to running the experiments, one agent with a 2D lidar was used to scan the environment and to create a map. This map was shared with all the other agents. All experiments where performed on a laptop with an AMD Ryzen 7 PRO 5850U, 32 GB RAM and Ubuntu 20.04 as the operating system. The proposed framework was implemented using C++ programming within ROS. The behavior trees are created and executed using the framework BehaviorTree.CPP \cite{bt_cpp_git}, the optimization problem associated with the auctioning system is solved using SCIP \cite{GamrathEtal2020OO} and a Vicon MoCap system was used to get the odometry of each TurtleBot.

\subsection{Evaluation Scenario 1}
The first scenario consists of one ground agent that is moving to pick up an object and deliver it to a specified location. At a random time, an "inspection" task is introduced and, depending on the stage of execution of the current task, the architecture has to reactively reallocate the available tasks.

\begin{figure}
    \centering
    \begin{subfigure}[t]{0.325\columnwidth}
        \centering
        \includegraphics[width=\textwidth]{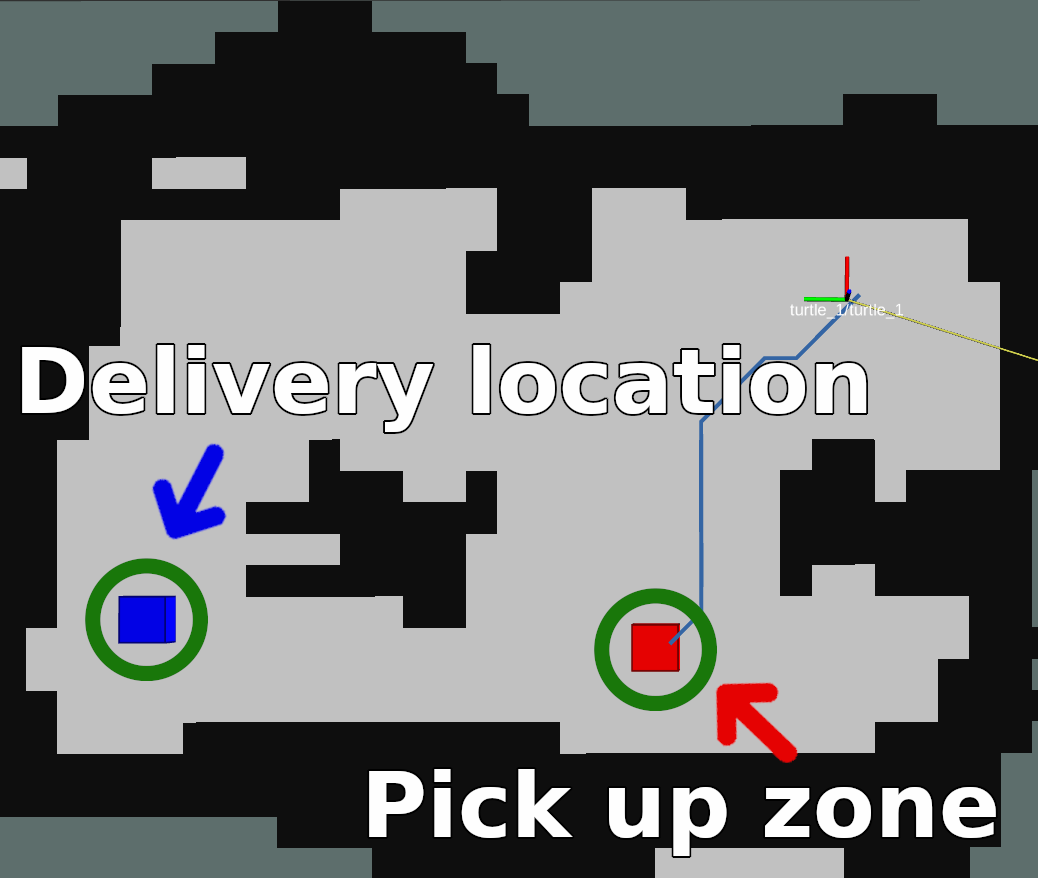}
        \caption{\(t = 0\text{ s}\).}
        \label{fig:evaluation_scenario_1_snapshots_a}
    \end{subfigure}
    \hfill
    \begin{subfigure}[t]{0.325\columnwidth}
        \centering
        \includegraphics[width=\textwidth]{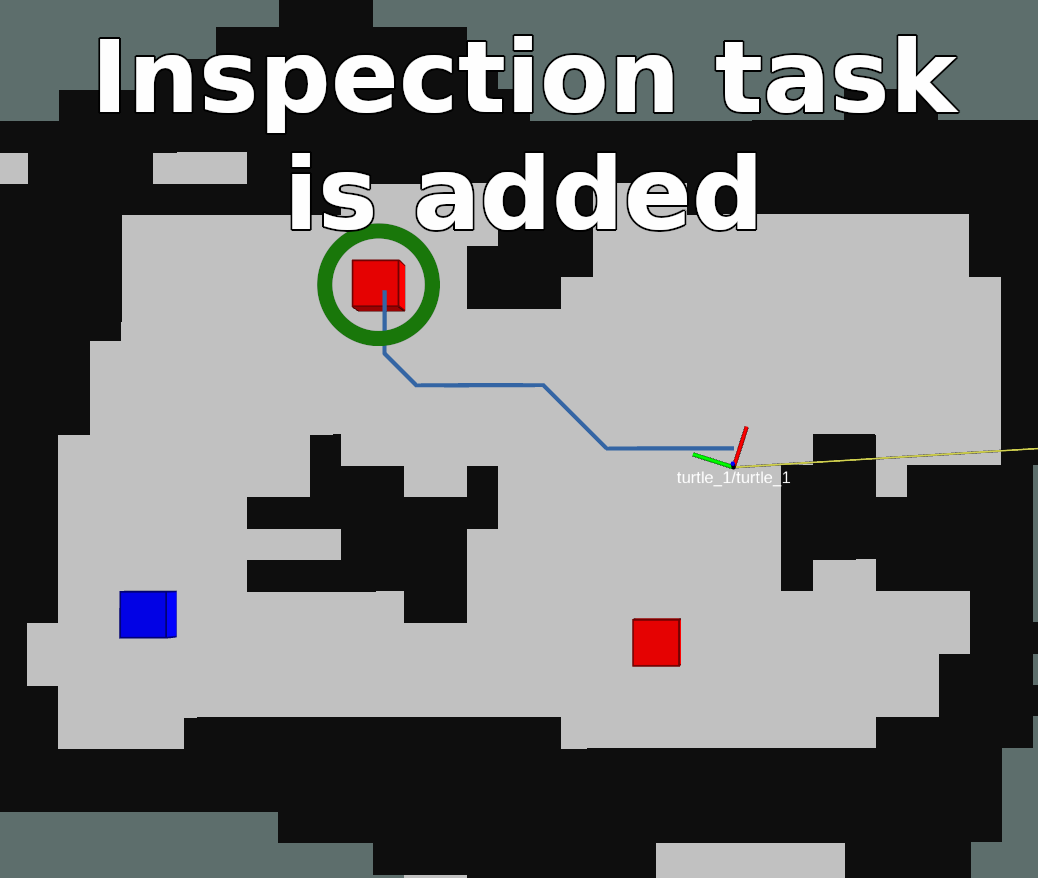}
        \caption{\(t = 6\text{ s}\)}
        \label{fig:evaluation_scenario_1_snapshots_b}
    \end{subfigure}
    \hfill
    \begin{subfigure}[t]{0.325\columnwidth}
        \centering
        \includegraphics[width=\textwidth]{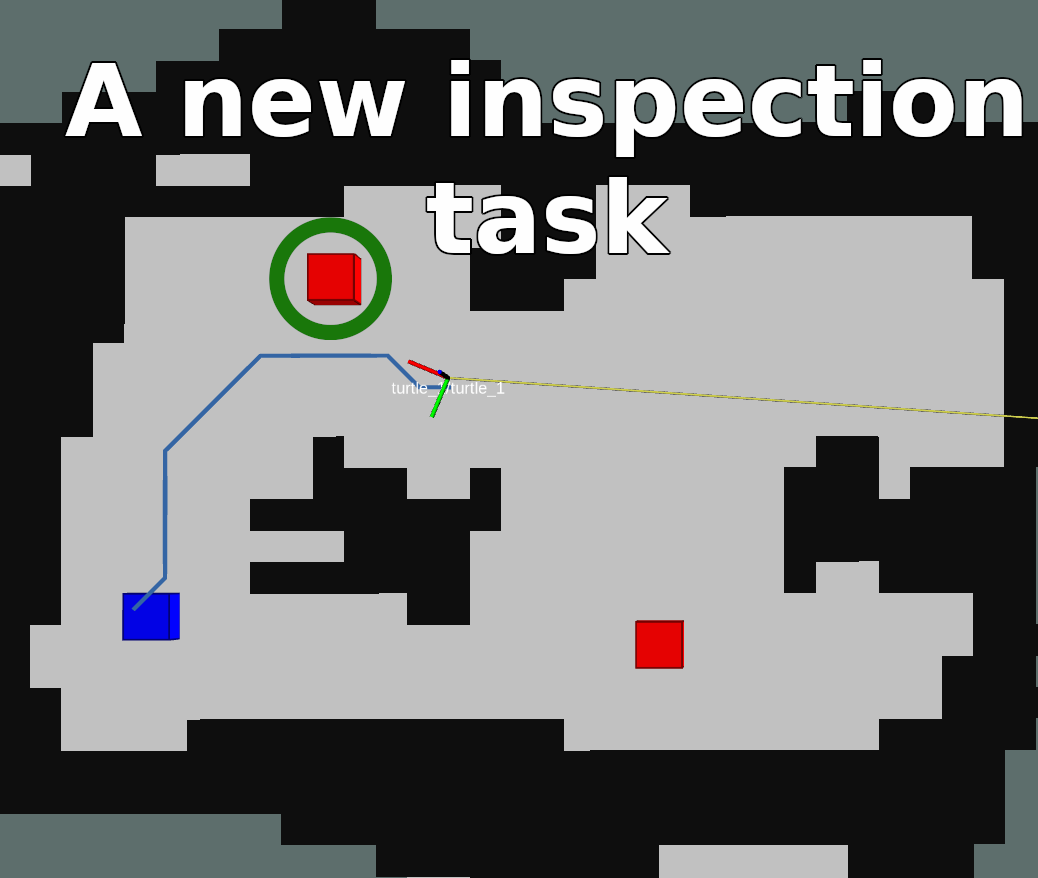}
        \caption{\(t = 40\text{ s}\)}
        \label{fig:evaluation_scenario_1_snapshots_c}
    \end{subfigure}

    \caption{Three snapshots demonstrating how the proposed architecture reallocates tasks, depending on the state of execution of the current task. (a) shows how the agent is allocated to a "pick and deliver" task, (b) how an "inspection" task is added and the agent reactively switches tasks. In (c) another inspection task is added after the agent finished the first stage of the "pick and deliver" task. Now there is no reallocation since the agent is not allowed to abandon its current task after picking up an object.
    }
    \label{fig:evaluation_scenario_1_snapshots}
\end{figure}

The results of the first evaluation case are shown in Fig \ref{fig:evaluation_scenario_1_snapshots}, demonstrating the ability of the proposed architecture to reactively allocate tasks depending on the current local state of the individual agents. Fig. \ref{fig:evaluation_scenario_1_snapshots_a} shows how the agent is allocated to a "pick and deliver" task. Fig. \ref{fig:evaluation_scenario_1_snapshots_b} shows how an "inspection" task is added and the agent reactively switches to the "inspection" task. In Fig. \ref{fig:evaluation_scenario_1_snapshots_c} an inspection task is added after the agent finished the first "pick" stage of the "pick and deliver" task. We see that here there is no reallocation since the architecture does not allow the agent to abandon its current task after picking up an object.



\subsection{Evaluation Scenario 2}

The second evaluation scenario consists of a warehouse inspired setup shown in Fig. \ref{fig:lab_example}, where multiple TurtleBots are available for completing several two-stage tasks involving going to a position, picking up an object and then dropping it at a delivery location. In this scenario, the tasks are introduced at random times, (their arrival times are not known a priori), with associated random pick-up and delivery locations. We demonstrate that the TurtleBots reactively work together as a team to complete all available tasks.


Between two successive implementation of the auctioning system, a delay of \(100\) ms was added at the end of the Announcement stage to set a reasonable update rate of the auction system, as discussed previously in section \ref{numerical_study}. The resulting update rate was found to be \(9\) Hz, that is, the architecture reactively incorporated newly arrived tasks into the auctioning system within $\left(\frac{1}{9}\right)^{\text{th}}$ of a second after their arrival. For the most clear demonstration of the efficacy of the proposed architecture, the reader is recommended to watch the video of the experiment at: \href{https://youtu.be/ZdEkoOOlB2g}{https://youtu.be/ZdEkoOOlB2g}. 


%
\begin{figure}
    \centering
    \begin{subfigure}[t]{0.325\columnwidth}
        \centering
        \includegraphics[width=\textwidth]{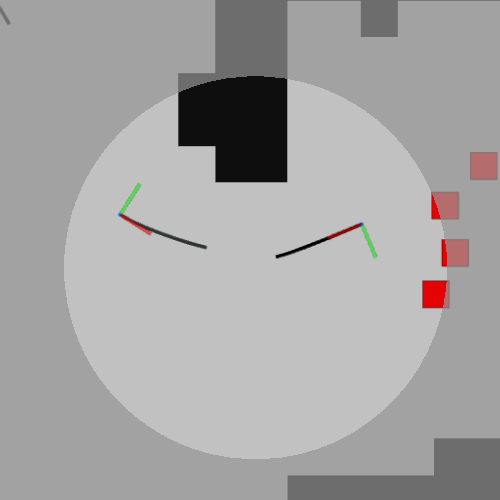}
        \caption{\(t = 66\text{ s}\).}
        \label{fig:collisions_snapshots_a}
    \end{subfigure}
    \hfill
    \begin{subfigure}[t]{0.325\columnwidth}
        \centering
        \includegraphics[width=\textwidth]{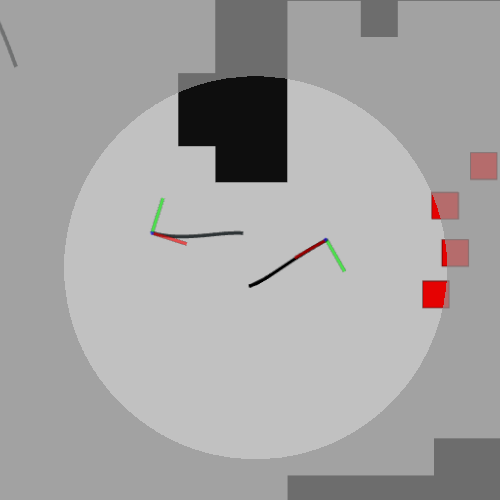}
        \caption{\(t = 67\text{ s}\)}
        \label{fig:collisions_snapshots_b}
    \end{subfigure}
    \hfill
    \begin{subfigure}[t]{0.325\columnwidth}
        \centering
        \includegraphics[width=\textwidth]{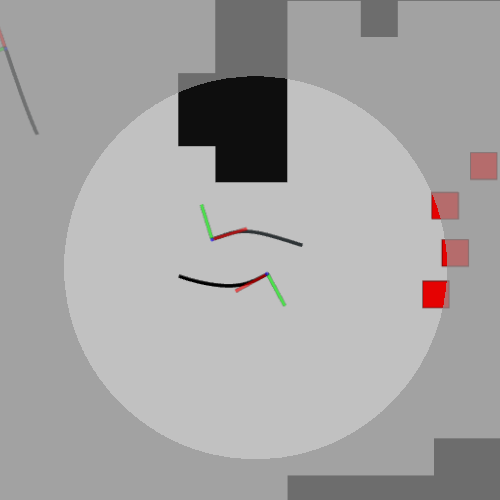}
        \caption{\(t = 70\text{ s}\)}
        \label{fig:collisions_snapshots_c}
    \end{subfigure}

    \caption{Three snapshots showing the NMPC-based distributed collision avoidance during a collision scenario, the black lines shows how the individual agents are predicting their future movement.}
    \label{fig:collisions_snapshots}
\end{figure}

\begin{figure}[ht!]
	\centering
	\includegraphics[width=0.99\columnwidth]{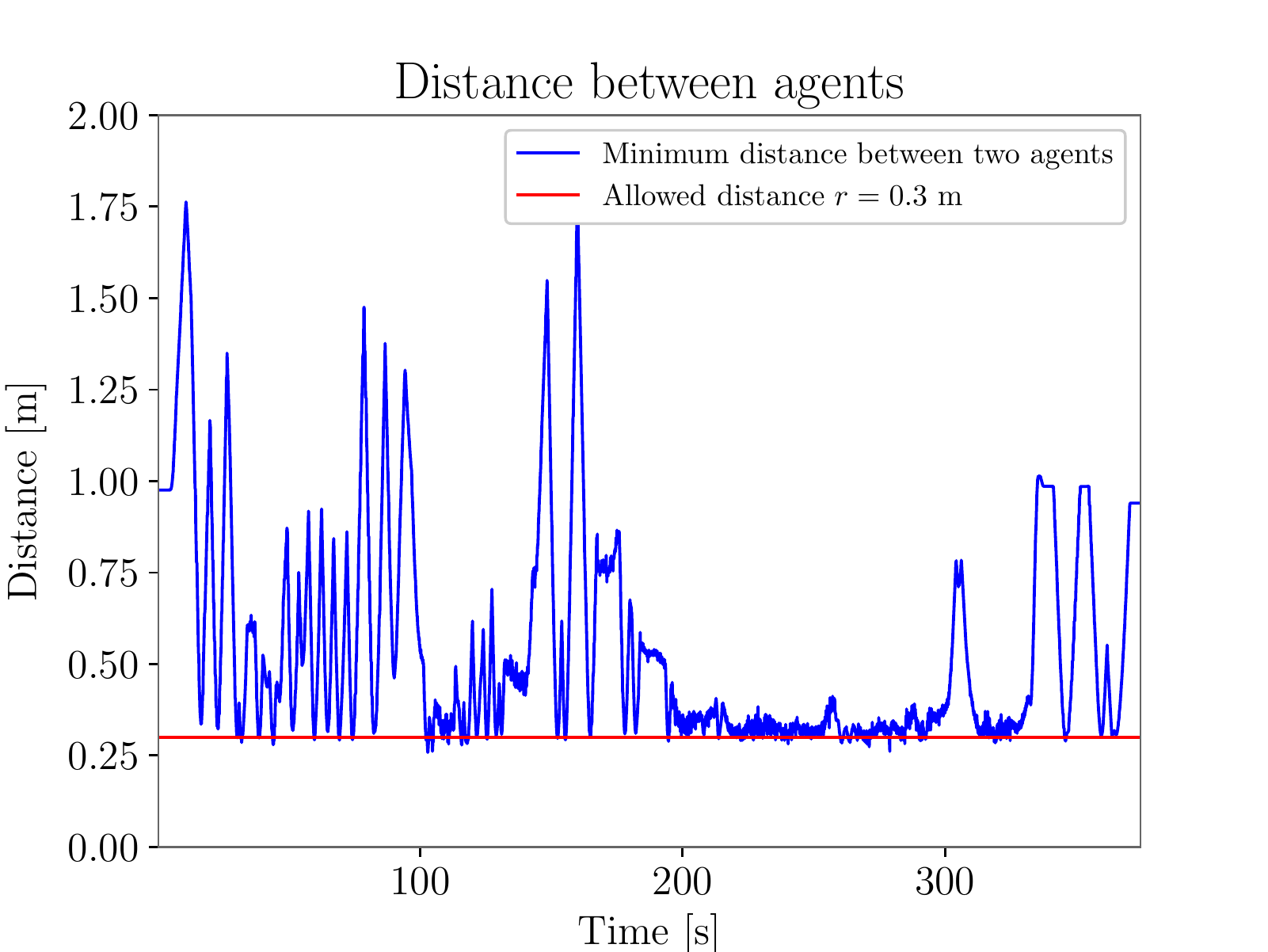}
  	\caption{Minimum agent-agent safety distances throughout the second scenario experiment. The safety radius is set to $\unit[0.3]{m}$ and plotted as a red line.}
  	\label{fig:dist}
\end{figure}

The "pick and deliver" scenario generated multiple instances, where agents are acting in close proximity and are often on direct collision courses with each other due to the shared drop-off location and "delivery lanes". Throughout this experimental scenario, the minimum distance between two agents is plotted in Fig \ref{fig:dist}. Based on the size of the TurtleBots, we select a desired clearance between agents $r_\mathrm{obs}$ to be $\unit[0.3]{m}$ and we see that this is maintained, with a maximum violation of $\unit[0.04]{m}$ at one instant. The collisions were cleanly avoided at all points in the experiment. Fig. \ref{fig:dist} and the video of the experiment show the performance of the NMPC module. Additionally, Fig. \ref{fig:collisions_snapshots} shows three snapshots showing how the agents predict their future positions during a collision scenario and successfully avoid collisions using NMPC.



\subsection{Evaluation Scenario 3}
The third and last evaluation scenario consists of a team of agents that are requested to do a number of "inspection" tasks and during execution, a number of high priority "inspection" tasks are introduced to demonstrate the reallocation that occurs. The prioritization is included in the optimization problem \eqref{eq:auction_optimization} by adding a large scaling constant to each profit related to the high priority tasks, as hinted in Section \ref{sec:auction_ilp}. 
\begin{figure}[htbp]
    \centering
    \begin{subfigure}[t]{0.525\columnwidth}
        \centering
        \includegraphics[width=\textwidth]{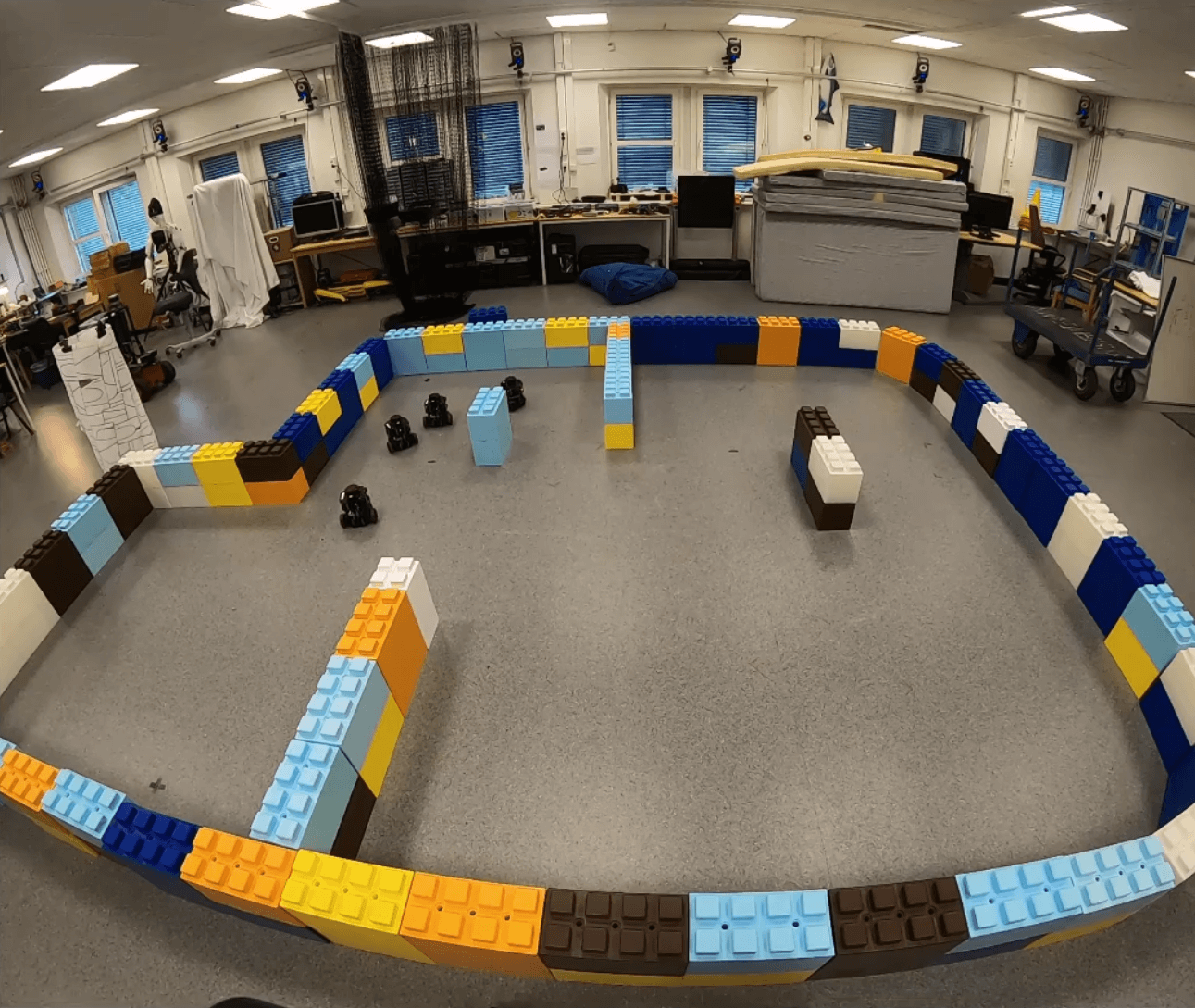}
        \caption{}
    \end{subfigure}
    \hfill
    \begin{subfigure}[t]{0.455\columnwidth}
        \centering
        \includegraphics[width=\textwidth]{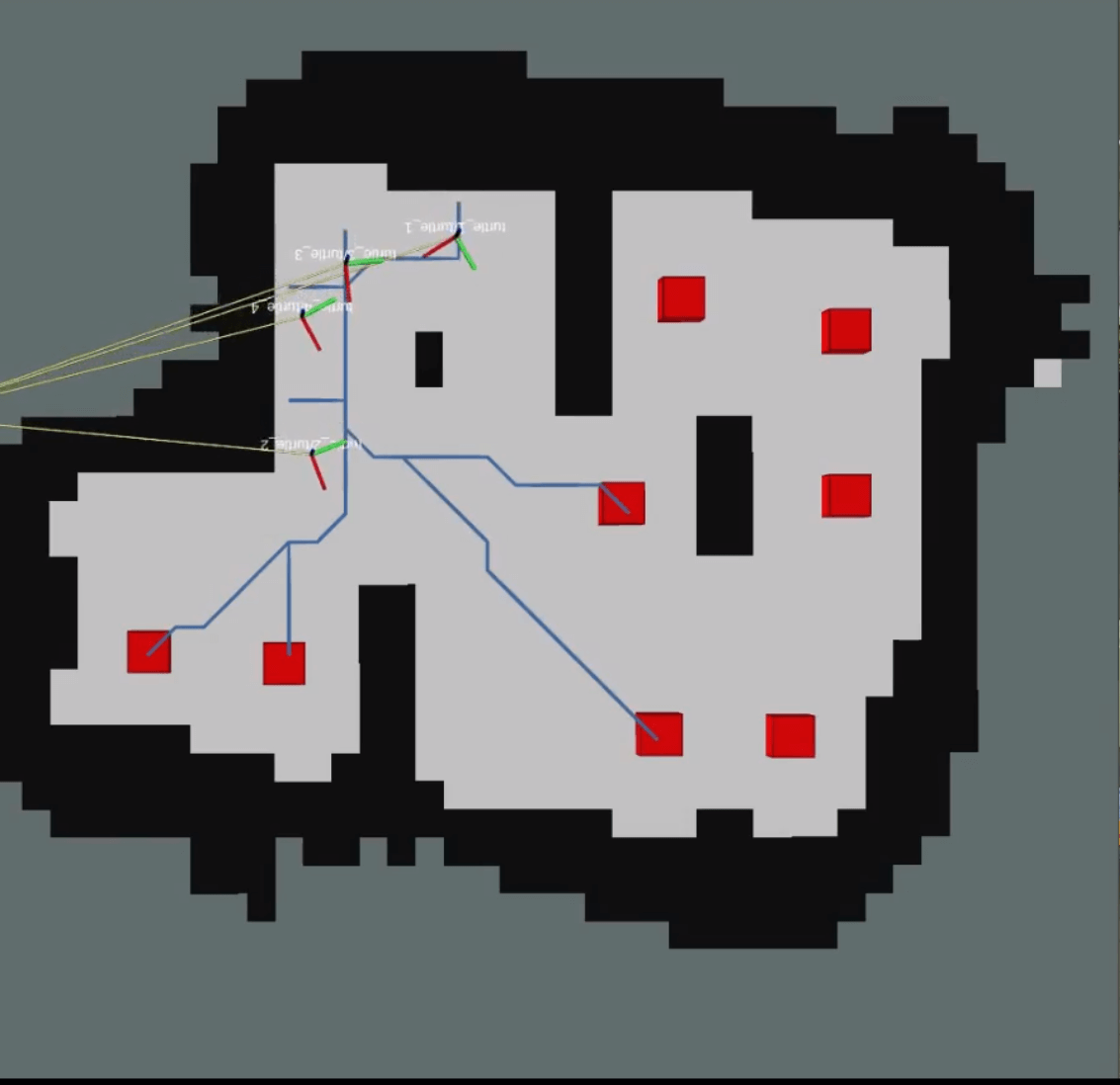}
        \caption{}
    \end{subfigure}
    \caption{In (a), the lab setting where scenario 3 was evaluated is shown. In (b) the static map used for path planning and cost estimation is shown, with red cubes indicating the initial "inspection" tasks.} 
    \label{fig:eval_3_tasks}
\end{figure}

\begin{figure}[htbp]
    \centering
    \begin{subfigure}[t]{0.325\columnwidth}
        \centering
        \includegraphics[width=\textwidth]{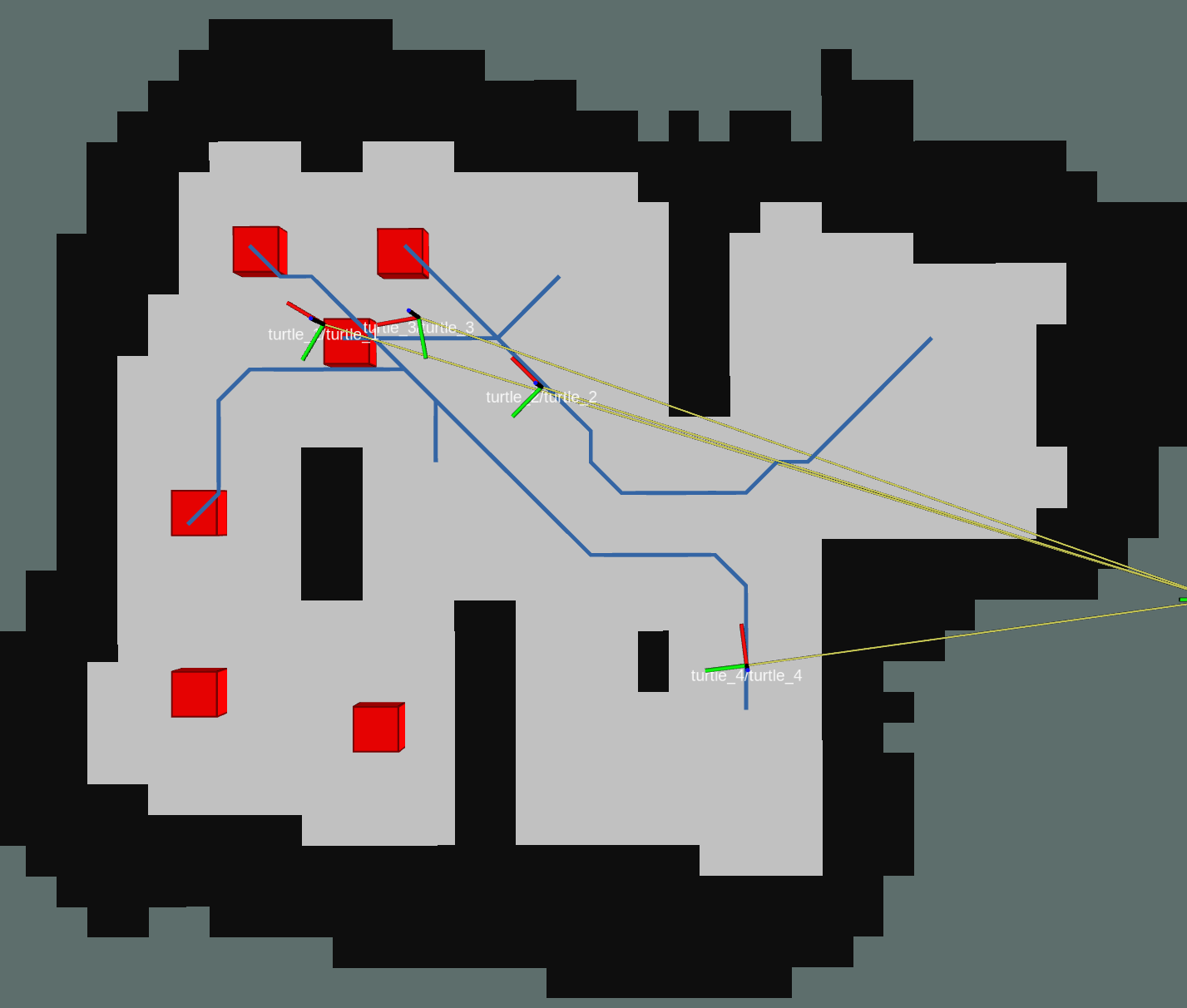}
        \caption{\(t = 27\text{ s}\).}
        \label{fig:evaluation_scenario_3_snapshots_a}
    \end{subfigure}
    \hfill
    \begin{subfigure}[t]{0.325\columnwidth}
        \centering
        \includegraphics[width=\textwidth]{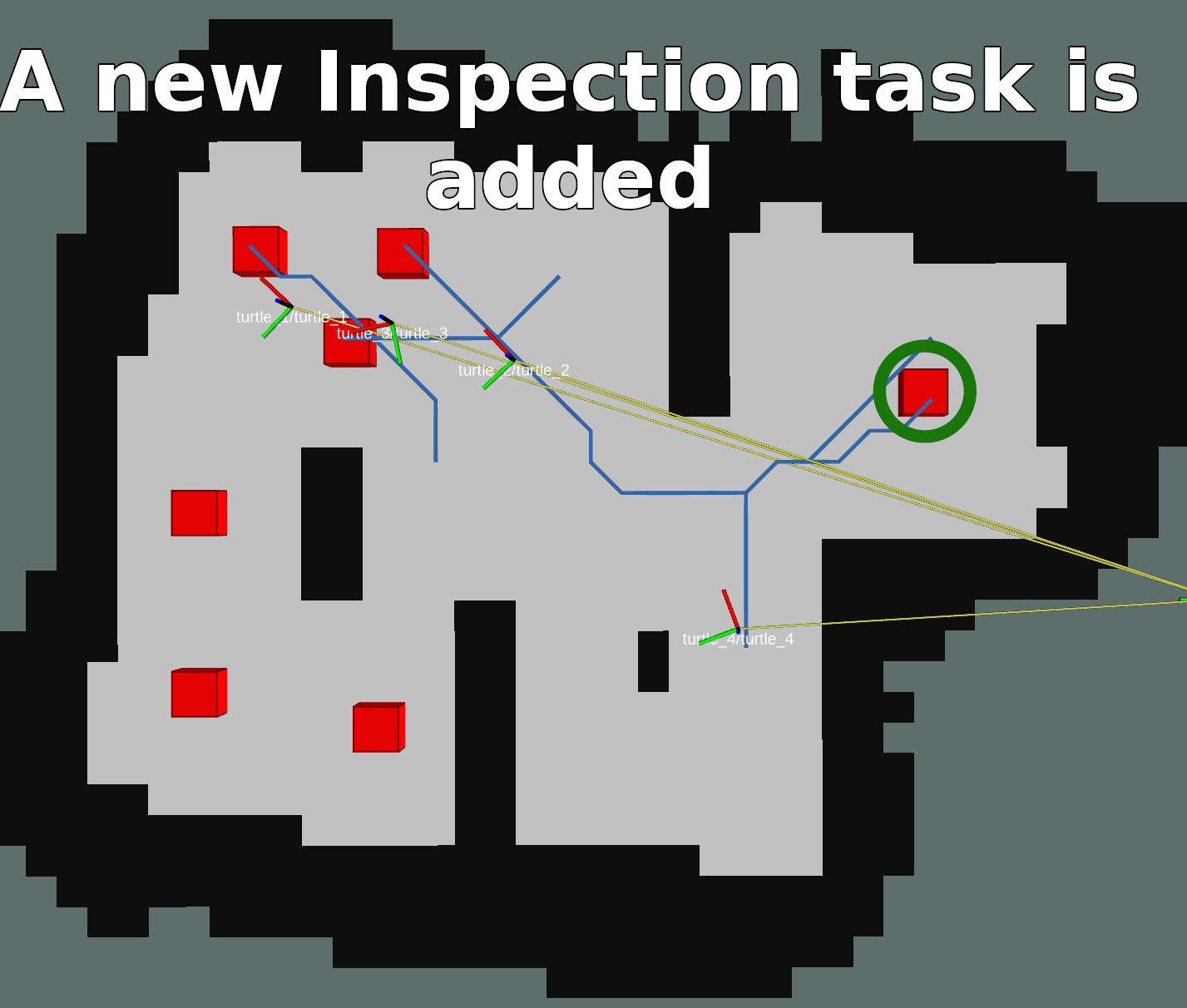}
        \caption{\(t = 28\text{ s}\)}
        \label{fig:evaluation_scenario_3_snapshots_b}
    \end{subfigure}
    \hfill
    \begin{subfigure}[t]{0.325\columnwidth}
        \centering
        \includegraphics[width=\textwidth]{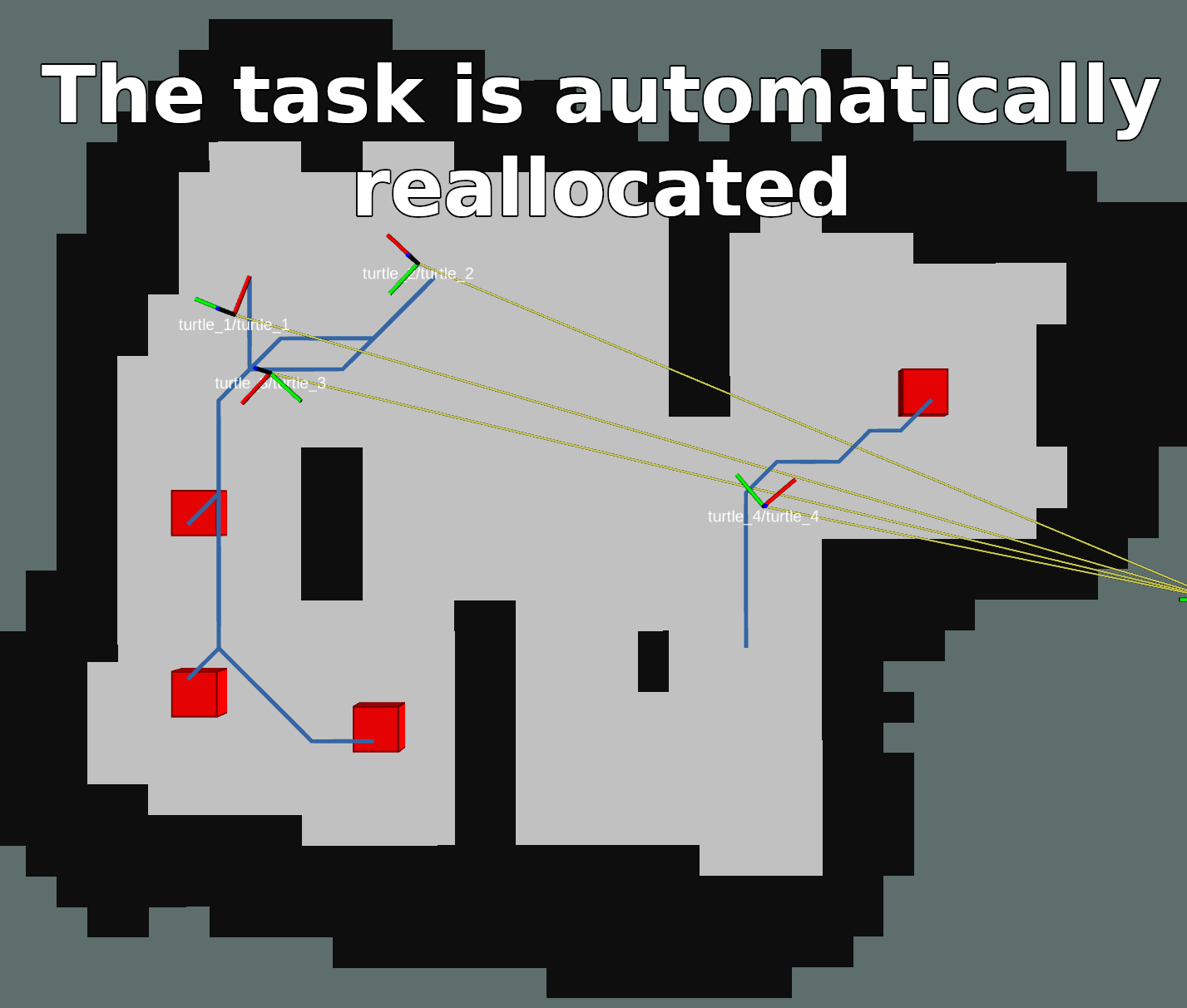}
        \caption{\(t = 32\text{ s}\)}
        \label{fig:evaluation_scenario_3_snapshots_c}
    \end{subfigure}

    \caption{Three snapshots demonstrating how the proposed architecture reallocates tasks during an inspection mission. In (b), a new high-priority "inspection" task is added, highlighted with a green circle, and the new task is then automatically allocated to an agent as shown in (c).
    }
    \label{fig:evaluation_scenario_3_snapshots}
\end{figure}

In Fig. \ref{fig:eval_3_tasks}, we see a team of agents executing a number of "inspection" tasks initially assigned to them as part of the third evaluation scenario. The initial tasks are shown as red cubes and six additional tasks are added during execution. Fig. \ref{fig:evaluation_scenario_3_snapshots} shows three snapshots with Fig. \ref{fig:evaluation_scenario_3_snapshots_b} covering a key moment where a high-priority task is added at \(t = 28\text{ s}\). In Fig. \ref{fig:evaluation_scenario_3_snapshots_c} we see how the architecture reallocates the agents from their current tasks in response to the newly added high-priority task. This, in combination with the video linked next, displays that the agents are able to complete all available tasks and reactively handle newly added tasks. A video demonstrating the reallocation is provided at \href{https://drive.google.com/drive/folders/1RepKMHyYOVgHWAVltBiQv2elU7luVA2t?usp=sharing}{
https://drive.google.com/drive/folders/\newline1RepKMHyYOVgHWAVltBiQv2elU7luVA2t
}. 

\section{Conclusions}
This work proposed a novel architecture for multi-agent coordination that integrates an auction system for task allocation with behavior tree management to control the individual robots. The auction system gathers the costs for executing the available tasks from the agents and reactively solves a linear integer program to decide the task distribution among the agents. The agents locally estimated the costs of completing available tasks and this allows the centralized auction system to assign tasks, without full global knowledge of the agents. The costs computed by the agents incorporated the state of a local behavior tree, to take into account the current state of multi-stage task execution, and this influenced the auctioning system's reallocation in response to newly introduced tasks. Three evaluation scenarios demonstrated the key aspects of the proposed architecture in a realistic lab environment. Future work includes enhancing and generalizing the architecture to be able to include more complex tasks, include heterogeneous teams of robots where the robot type is taken into consideration in the allocation process, as well as inclusion of tasks where agents need to act in a cooperative manner to complete tasks.

\bibliography{mybib}

\end{document}